# An Event Grouping Based Algorithm for University Course Timetabling Problem

Velin Kralev, Radoslava Kraleva, Borislav Yurukov

Department of Informatics, South West University "Neofit Rilski", Blagoevgrad, Bulgaria

*Abstract* — **This paper presents the study of an event grouping based algorithm for a university course timetabling problem. Several publications which discuss the problem and some approaches for its solution are analyzed. The grouping of events in groups with an equal number of events in each group is not applicable to all input data sets. For this reason, a universal approach to all possible groupings of events in commensurate in size groups is proposed here. Also, an implementation of an algorithm based on this approach is presented. The methodology, conditions and the objectives of the experiment are described. The experimental results are analyzed and the ensuing conclusions are stated. The future guidelines for further research are formulated.**

## I. INTRODUCTION

The University Course Timetabling Problem (UCTP) is an optimization problem and has been widely explored for the last 55 years. For the first time the key aspects of this problem were presented in [1]. In order to solve a UCTP a finite number of events $E = \{e_1, e_2, ..., e_n\}$ synchronized in time and fixed on a timetable that consists of a finite number of time slots $T = \{t_1, t_2, ..., t_k\}$ is needed. The arrangement of the events must be done in such a way that it satisfies the finite number of hard constraints ($C_h$) and violates the fewest possible ones from a finite number of soft constraints ($C_s$). A timetable is acceptable when it meets all hard constraints and is better than another one when it violates fewer soft constraints [2].

The UCTP is NP-hard [3], but it has been intensively studied because of its great practical relevance [4], [5] and others. In recent years, the interest in the heuristic and hybrid approaches towards solving this problem has increased. These approaches give better results than the approaches based on constructive heuristics [6], [7] and [8].

There are different approaches that are used to solve the UCTP, for instance: constructive heuristics, meta-heuristics and constraints-based approaches. They are discussed in detail in the scientific literature [4], [9], [10], [11] and [12]. In addition to these approaches others are well known as well, for instance: multicriteria approaches, case-based reasoning, knowledge-based approaches and hyper-heuristic approaches [13].

### A. Constraint-based approaches

In addition to the use of constraints in the constraint-based approaches, other supporting methods are used, such as: "Depth First Search", object-oriented modeling of graphs and trees, "backtracking", combined methods and genetic algorithms [14]. The experimental results show that it is possible for certain acceptable time to find good solutions that are close to the optimal one, but it refers only to timetables with a small number of events. This can be done by not considering temporary solutions that are not promising.

### B. Graph-based approaches

Graph-based approaches show how the UCTP can be represented by a graph [4]. The graph coloring problem and its relationship with the UCTP are widely discussed in the scientific literature, for instance in [15].

### C. Meta-heuristic and hyper-heuristic approaches

Meta-heuristic and hyper-heuristic approaches are methods of high level which are used to find the solution to problems with a large computational complexity. For instance, such are: "tabu search" [16]; "simulated annealing" [17]; "variable neighborhood search" [18] and "ant colony optimization" [19].

The purpose of these approaches is maximum satisfaction of the soft constraints. They are one of the most effective strategies for the practical solution to optimization problems. The published results indicate that the proposed methods find good solutions when they are used for UCTP. Their disadvantage is the need to set up additional parameters that control the performance of the algorithms.

### D. Case-based reasoning and knowledge-based approaches

Case-based reasoning approaches (CBR) are characterized by the fact that additional heuristic methods are used. For instance, graphs in which the attributes of the vertices and the edges store more information about the interconnection between events. In this way, the algorithm that generates a timetable shall decide how to continue the process from here (or to improve the final solution) [12] and [13]. Knowledge-based approaches use an expert system of rules with pre-defined strategies (for instance, "Depth-first search" [20].

*E. Population-based approaches*

In solving UCTP quite often population-based approaches are used. The most commonly used algorithms of this type are genetic and memetic [21] and their modifications in [22]. The published results indicate that these approaches generate good acceptable solutions for a short time.

An analytical description of the real UCTP is presented in [23]. The proposed model includes parameters, vectors and matrices, which are used in solving the problem, as well as a function to evaluate the found solutions. The soft constraints are described by weights which provides greater flexibility in their analysis. The implementation of a genetic algorithm (GA) and a memetic algorithm (MA), as well as their computational complexity (respectively, quadratic for GA and cubic for MA), are presented in [24]. These algorithms are used to solve the real UCTP. The solutions found are evaluated according to the model presented in [23]. It is shown experimentally that for the same input data GA generates good solutions comparable to those obtained by solving the problem of the user – expert. Unlike GA, MA generates better solutions (for all test input data sets) but runs slower because of its higher computing complexity [24].

In [25] an approach in which the events are grouped in groups of the same size is used. Then, the best solution to a given order of the events in the first group is looked for. Similarly, the best solutions in the order of events in the other groups are looked for. In this way the best solution for a given group cannot be worse than the last best solution found for the previous group. The results obtained for some input data are the best ones found so far. For the other tested input data sets, the algorithm found solutions commensurate with those found by MA [24]. However, not all possible groupings of events have been investigated (and only a small number of multiples of the number of events) which motivated the authors to focus on this subject of study in this article.

## II. An Event Grouping Based Algorithm

An event grouping based algorithm (EGB) to UCTP will be presented. All possible groupings of events in commensurate in size groups will be generated. The algorithm will search for the best solution for each successive order of events in each of the groups. It is necessary to determine how the number of groups affects the quality of the solutions found. As mentioned above (and described in [24]), for large size of the input data (on the order of several thousand events), the performance of the MA will take more computing time (due to the fact that more solutions should be found) in comparison with the algorithm, EGB which also will use the evaluation model presented in [23]. This algorithm is integrated into the updated version of the information system for the automated university course timetabling presented in [26].

Let $N$ is a set of $n$ events, i.e. $E = \{e_1, e_2, ..., e_n\}$, $n \geq 4$ and $G$ is a set of $m$ different ways of grouping these events, i.e. $G = \{g_1, g_2, ..., g_m\}$ such that $2 \leq m \leq \lfloor n/2 \rfloor$, or in other words it is necessary to establish at least two groups, as in any group, there are at least two events. The union of all groups of events gives the set $E$, i.e. $g_1 \cup g_2 \cup ... \cup g_m = E$, or in other words every event is in exactly one group, i.e. $g_i \cap g_j = \varnothing$, for $\forall\ i \neq j$. The cardinality of any two groups should not differ with more than one event, i.e., it must be satisfied:

$$\left\| |g_i| - |g_j| \right\| = \begin{cases} 0, & if\ (n \bmod m) = 0 \\ 1, & otherwise \end{cases},\quad i \neq j \qquad (1)$$

To satisfy (1) it is necessary that the $n \bmod m$ groups (i.e. the remainder of dividing the $n$ and $m$) have exactly the $\lfloor n/m \rfloor + 1$ events (i.e. the quotient of the division of $n$ and $m$ without remainder). Some other interesting techniques using grouping of resources (not necessarily the events) are found in the scientific literature, for example in [27] and [28].

Below an example with 11 events and their distribution in 2, 3, 4 and 5 groups is presented.

TABLE I.    Distribution of 11 Events into 2, 3, 4 and 5 Groups

| $m = 2$; $\lfloor n/m \rfloor = 5$; $(n \bmod m) = 1$; $\lfloor n/m \rfloor + 1 = 6$ | | | | | | | | | | |
|---|---|---|---|---|---|---|---|---|---|---|
| $e_n$ | $e_1$ | $e_2$ | $e_3$ | $e_4$ | $e_5$ | $e_6$ | $e_7$ | $e_8$ | $e_9$ | $e_{10}$ | $e_{11}$ |
| $g_2$ | $|g_1| = 6$ | | | | | | $|g_2| = 5$ | | | | |

| $m = 3$; $\lfloor n/m \rfloor = 3$; $(n \bmod m) = 2$; $\lfloor n/m \rfloor + 1 = 4$ | | | | | | | | | | |
|---|---|---|---|---|---|---|---|---|---|---|
| $e_n$ | $e_1$ | $e_2$ | $e_3$ | $e_4$ | $e_5$ | $e_6$ | $e_7$ | $e_8$ | $e_9$ | $e_{10}$ | $e_{11}$ |
| $g_3$ | $|g_1| = 4$ | | | | $|g_2| = 4$ | | | | $|g_3| = 3$ | | |

| $m = 4$; $\lfloor n/m \rfloor = 2$; $(n \bmod m) = 3$; $\lfloor n/m \rfloor + 1 = 3$ | | | | | | | | | | |
|---|---|---|---|---|---|---|---|---|---|---|
| $e_n$ | $e_1$ | $e_2$ | $e_3$ | $e_4$ | $e_5$ | $e_6$ | $e_7$ | $e_8$ | $e_9$ | $e_{10}$ | $e_{11}$ |
| $g_4$ | $|g_1| = 3$ | | | $|g_2| = 3$ | | | $|g_3| = 3$ | | | $|g_4| = 2$ | |

| $m = 5$; $\lfloor n/m \rfloor = 2$; $(n \bmod m) = 1$; $\lfloor n/m \rfloor + 1 = 3$ | | | | | | | | | | |
|---|---|---|---|---|---|---|---|---|---|---|
| $e_n$ | $e_1$ | $e_2$ | $e_3$ | $e_4$ | $e_5$ | $e_6$ | $e_7$ | $e_8$ | $e_9$ | $e_{10}$ | $e_{11}$ |
| $g_5$ | $|g_1| = 3$ | | | $|g_2| = 2$ | | $|g_3| = 2$ | | $|g_4| = 2$ | | $|g_5| = 2$ | |

After conducting the experiments and analyzing the obtained results it was found that the best solutions are not always generated when events are distributed in regular groups.

An implementation of the EGB algorithm will be presented in the Object Pascal (Delphi) language.

```pascal
procedure EventGrouping(n: integer);
var
  m, g, r: integer;
  tg, tr, tn, tm: integer;
  flag: boolean;
  i, j, count: integer;
  from_index, to_index, best_index: integer;
  first, tmp: integer;
  eval, best_eval: single;
  p: array of integer;
  e: array of integer; //an array of events
  groups: array of integer; //an array of groups
  col, row: integer;
begin
```

```
setlength(p, n); //memory allocation for p
setlength(e, n); //memory allocation for e
for m := 2 to (n div 2) do //for each group
begin
   g := n div m; //events in group
   r := n mod m; //undistributed events
   groups := nil; //deallocate groups array
   setlength(groups, m + 1); //allocate memory
   tg := g; //number of events
   tr := r; //undistributed events
   flag := false; //a boolean variable
   count := 0;
   tm := 1; //the first group
   groups.cells[1, 1] := 1;
   for tn := 1 to n do
   begin
      e.cells[tn, 1] := tm;
      tg := tg - 1; //an event is fixed
      count := count + 1; //the same as inc(count)
      if ((tg = 0) and (tr > 0) and (not flag)) then
      begin
         tg := 1;
         tr := tr - 1; //the same as dec(tr)
         flag := true;
         continue; //continue to the next iteration
      end;
      if (tg = 0) then
      begin
         groups.cells[2, tm] := tn;
         groups.cells[3, tm] := count;
         tm := tm + 1; //the same as inc(tm)
         tg := g;
         count := 0;
         if (tr > 0) then flag := false;
         if (tm <= m) then
            groups.cells[1, tm] := tn + 1;
      end;
   end; //for tn := 1 to n do
   for tm := 1 to m do //for each group
   begin
      from_index := groups.cells[1, tm];
      to_index := groups.cells[2, tm];
      best_eval := maxint; //init best_eval
      best_index := 0; //init best_index
      for i := from_index to to_index do
      begin
         LocalSearch; //call LocalSearch method
         if (eval < best_eval) then
         begin
            best_eval := eval;
            best_index := i;
         end;
         //move events from from_index to to_index
         //to the left one position
         first := p[from_index];
         for j := from_index to to_index - 1 do
            p[j] := p[j + 1];
         p[to_index] := first;
      end; //for i := from_index to to_index do
      tmp := p[1];
      for j := 1 to (best_index - from_index) do
         p[j] := p[j + 1];
      p[j] := tmp;
   end; //for tm := 1 to m do
end; //for m := 2 to (n div 2) do
end; //end EventGrouping method
```

For each grouping *m* the EGB algorithm rearranges all events *n*. After each rearrangement of the events (in a group) the local search method is called which finds the best solution in this order of events. As the complexity of the *LocalSearch* method is the quadratic [24], for the proposed algorithm it is found out that there is a computational complexity $O = m.n^3$. In the General case the complexity is cubic which also depends on the number of groupings $m = n/2 − 1$. Finding a way to reduce the number of groupings will reduce the execution time of the EGB algorithm.

## III. EXPERIMENTAL RESULTS

The object of the study is an updated version of the integrated information system to university course timetabling. Its development and use are described in [26]. In the updated version of the system and EGB algorithm, that was presented above, was added (Fig. 1).

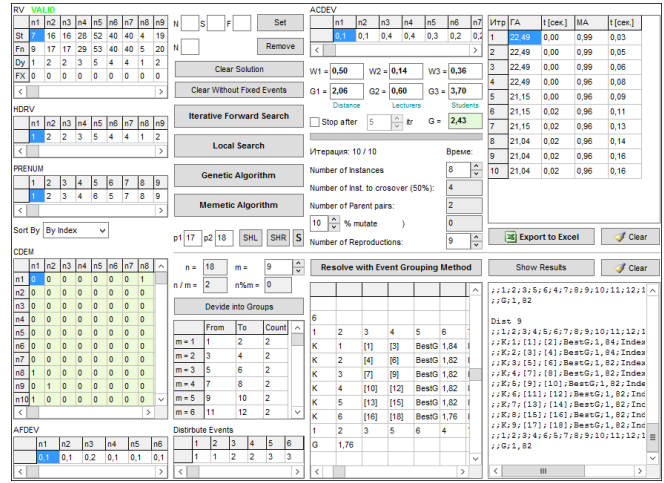

Figure 1. Working session with the updated version of the system.

With this system specific experiments to test the EGB algorithm with real data can be made.

The aim of the experiments was to determine the behavior of the algorithm on specific input data sets which are presented in [24]. For these input data sets there is already information concerning the algorithms used and the best solutions found. For some input data the EGB algorithm generated the best currently known solutions so far. In order to determine (experimentally) under what groupings of events the best results are received, all possible groupings will be generated.

### A. Experimental Conditions

The experimental conditions for conducting the experiments are the following: PC with 64-bit Operating System Windows 10 Pro, x64-based processor and the following hardware configuration: Processor: Intel(R) Core(TM) i7-4712MQ CPU at 2.30 GHz; RAM memory: 8 GB DDR3 L.

### B. Methodology of the experiment

To achieve the goals of the experiments three input data sets were used:

- Input data set DS_E90S175L29A18 with ninety events (90), one hundred and seventy-five students (175), twenty-nine lecturers (29) and eighteen auditoriums (18);

- Input data set DS_E130S274L37A22 with one hundred and thirty events (130), two hundred and seventy-four students (274), thirty-seven lecturers (37) and twenty-two auditoriums (22);

- Input data set DS_E273S549L62A39 with two hundred and seventy-three events (273), five hundred and forty-nine students (549), sixty-two lecturers (62) and thirty-nine auditoriums (39).

### C. Experimental results

In Fig. II, the results of the EGB algorithm execution on input data set DS_E90S175L29A18 are shown. The events are sorted in order by index, weight, number and duration. This sequence was the same in all experiments.

TABLE II.   RESULTS FOR DS_E90S175L29A18

| m | Groups | Index | Weight | Number | Duration |
|---|--------|-------|--------|--------|----------|
| 2 | 2x45 | 9.758 | 7.422 | 8.545 | 6.967 |
| 3 | 3x30 | 9.312 | 6.530 | 7.453 | 8.002 |
| 4 | 2x23; 2x22 | 9.120 | 7.652 | 7.198 | 7.835 |
| 5 | 5x18 | 7.304 | 6.817 | 7.821 | 7.695 |
| 6 | 6x15 | 7.561 | 6.597 | 6.823 | 7.137 |
| 7 | 6x13; 1x12 | 7.618 | 7.469 | 8.207 | 7.487 |
| 8 | 2x12; 6x11 | 7.589 | 7.459 | 7.228 | 8.047 |
| 9 | 9x10 | 7.018 | 7.247 | 8.278 | 8.423 |
| 10 | 10x9 | 8.740 | 7.464 | 8.637 | 8.314 |
| 11 | 2x9; 9x8 | 9.365 | 7.491 | 8.604 | 8.418 |
| 12 | 6x8; 6x7 | 8.341 | 7.431 | 8.990 | 7.140 |
| 13 | 12x7; 1x6 | 7.598 | 7.502 | 6.759 | 7.264 |
| 14 | 6x7; 8x6 | 7.811 | 7.529 | 6.787 | 7.264 |
| 15 | 15x6 | 8.987 | 7.529 | 7.170 | 7.662 |
| 16 | 10x6; 6x5 | 9.107 | 7.518 | 7.170 | 7.922 |
| 17 | 5x6; 12x5 | 8.999 | 7.518 | 7.154 | 7.922 |
| 18 | 18x5 | 8.029 | 7.902 | 7.319 | 7.635 |
| 19 | 14x5; 5x4 | 8.085 | 7.902 | 7.319 | 7.647 |
| 20 | 10x5; 10x4 | 7.516 | 7.902 | 7.319 | 7.879 |
| 21 | 6x5; 15x4 | 8.879 | 7.902 | 7.319 | 7.879 |
| 22 | 2x5; 20x4 | 10.542 | 7.902 | 7.217 | 7.660 |
| 23 | 21x4; 2x3 | 9.936 | 7.718 | 8.882 | 7.791 |
| 24 | 18x4; 6x3 | 9.936 | 7.799 | 8.882 | 8.353 |
| 25 | 15x4; 10x3 | 9.931 | 7.853 | 8.905 | 8.353 |
| 26 | 12x4; 14x3 | 9.931 | 8.060 | 8.905 | 8.359 |
| 27 | 9x4; 18x3 | 9.931 | 8.060 | 8.905 | 8.359 |
| 28 | 6x4; 22x3 | 10.946 | 8.060 | 8.748 | 8.359 |
| 29 | 3x4; 26x3 | 10.946 | 8.060 | 9.924 | 8.359 |
| 30 | 30x3 | 8.371 | 8.421 | 8.349 | 9.090 |
| 31 | 28x3; 3x2 | 8.371 | 8.421 | 8.349 | 9.090 |
| 32 | 26x3; 6x2 | 8.371 | 8.421 | 8.349 | 9.090 |
| 33 | 24x3; 9x2 | 8.376 | 8.421 | 8.349 | 9.090 |
| 34 | 22x3; 12x2 | 8.376 | 8.421 | 8.366 | 9.090 |
| 35 | 20x3; 15x2 | 8.376 | 8.421 | 8.366 | 9.090 |
| 36 | 18x3; 18x2 | 8.376 | 8.421 | 8.366 | 9.090 |
| 37 | 16x3; 21x2 | 8.376 | 8.421 | 8.366 | 9.172 |
| 38 | 14x3; 24x2 | 8.387 | 8.719 | 8.366 | 9.172 |
| 39 | 12x3; 27x2 | 8.387 | 8.719 | 8.366 | 9.172 |
| 40 | 10x3; 30x2 | 8.387 | 8.719 | 8.361 | 9.172 |
| 41 | 8x3; 33x2 | 10.308 | 8.719 | 8.361 | 9.183 |
| 42 | 6x3; 36x2 | 10.308 | 8.719 | 8.361 | 7.504 |
| 43 | 4x3; 39x2 | 10.308 | 8.719 | 9.421 | 7.504 |
| 44 | 2x3; 42x2 | 11.100 | 8.071 | 9.700 | 8.194 |
| 45 | 45x2 | 11.100 | 8.769 | 9.239 | 8.194 |

The influence of the group number on the solution value for an input data set DS_E90S175L29A18 is shown in Fig. 2.

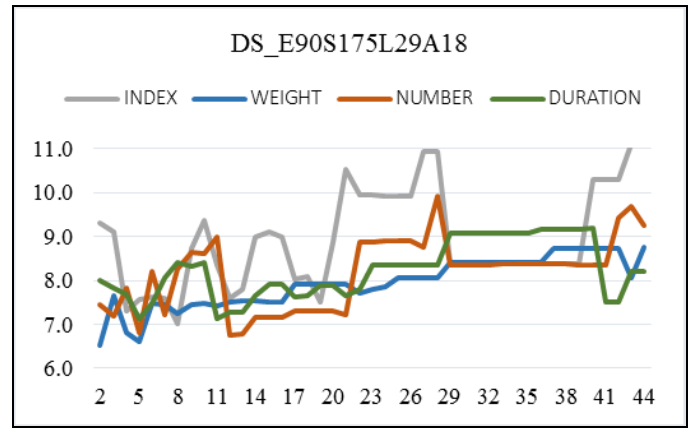

Figure 2.   Influence of the group number (the x axis) on the solution value (the y axis) for DS_E90S175L29A18.

In Fig. III, the best results of the EGB algorithm execution on an input data set DS_E90S175L29A18 (for each sort criteria) are shown.

TABLE III.   THE BEST RESULTS FOR DS_E90S175L29A18

| By | Index | Weight | Number | Duration |
|----|-------|--------|--------|----------|
| Best | m=9: 7.018 | m=3: 6.530 | m=13: 6.759 | m=2: 6.967 |

The influence of the sort criteria on the best solution value for an input data set DS_E90S175L29A18 is shown in Fig. 3.

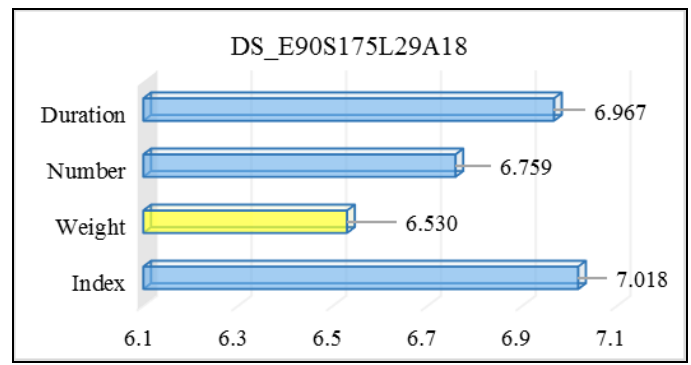

Figure 3.   Influence of the sort criteria (the y axis) on the best solution value (the x axis) for DS_E90S175L29A18.

Fig. II, III, 2 and 3 show that for the input data set DS_E90S175L29A18 the best found solution is with a value of 6.530. The solution was obtained when the events were sorted by weight and divided into 3 groups (respectively with 30 events in each). Another good solution (with a value of 6.759) was found when the events were sorted by number and divided into 13 groups (12 groups with 7 events and a group with 6 events). When the events were sorted by index, the best found solution (with a value of 7.018) is the worst found solution of all other solutions found when sorting the events in the other three criteria.

In Fig. IV, the results of the EGB algorithm execution on input data set DS_E130S274L37A22 are shown.

TABLE IV.    RESULTS FOR DS_E130S274L37A22

| m | Groups | Index | Weight | Number | Duration |
|---|--------|-------|--------|--------|----------|
| 2 | 2x65 | 12.227 | 11.489 | 11.331 | 11.547 |
| 3 | 1x44; 2x43 | 12.762 | 11.177 | 11.200 | 11.147 |
| 4 | 2x33; 2x32 | 15.118 | 10.170 | 11.332 | 10.556 |
| 5 | 5x26 | 12.476 | 9.689 | 11.328 | 9.707 |
| 6 | 4x22; 2x21 | 11.824 | 10.283 | 11.659 | 10.820 |
| 7 | 4x19; 3x18 | **10.070** | 10.006 | 11.331 | 10.526 |
| 8 | 2x17; 6x16 | 11.692 | **9.158** | 10.552 | 10.882 |
| 9 | 4x15; 5x14 | 10.580 | 9.677 | 11.864 | 11.663 |
| 10 | 10x13 | 10.714 | 10.070 | 10.470 | 10.623 |
| 11 | 9x12; 2x11 | 10.878 | 9.787 | 10.703 | 12.108 |
| 12 | 10x11; 2x10 | 12.170 | 10.000 | **10.239** | **8.958** |
| 13 | 13x10 | 13.422 | 10.032 | 11.155 | 9.549 |
| 14 | 4x10; 10x9 | 13.254 | 10.093 | 11.376 | 9.658 |
| 15 | 10x9; 5x8 | 10.423 | 9.509 | 11.057 | 10.236 |
| 16 | 2x9; 14x8 | 11.306 | 9.628 | 12.298 | 10.685 |
| 17 | 11x8; 6x7 | 10.823 | 10.286 | 10.954 | 10.867 |
| 18 | 4x8; 14x7 | 11.683 | 10.297 | 11.491 | 11.592 |
| 19 | 16x7; 3x6 | 13.353 | 10.774 | 12.272 | 9.268 |
| 20 | 10x7; 10x6 | 13.797 | 10.774 | 12.299 | 9.307 |
| 21 | 4x7; 17x6 | 13.797 | 10.918 | 12.299 | 9.716 |
| 22 | 20x6; 2x5 | 12.620 | 11.130 | 11.339 | 10.035 |
| 23 | 15x6; 8x5 | 12.844 | 11.122 | 11.347 | 10.020 |
| 24 | 10x6; 14x5 | 12.214 | 10.842 | 11.347 | 10.020 |
| 25 | 5x6; 20x5 | 13.225 | 10.386 | 11.604 | 10.514 |
| 26 | 26x5 | 13.729 | 10.473 | 11.524 | 11.737 |
| 27 | 22x5; 5x4 | 13.729 | 10.481 | 11.535 | 11.737 |
| 28 | 18x5; 10x4 | 13.729 | 10.481 | 11.539 | 11.741 |
| 29 | 14x5; 15x4 | 13.800 | 10.893 | 11.539 | 11.752 |
| 30 | 10x5; 20x4 | 13.800 | 11.008 | 11.539 | 11.752 |
| 31 | 6x5; 25x4 | 15.548 | 11.148 | 11.651 | 11.878 |
| 32 | 2x5; 30x4 | 11.828 | 11.162 | 11.651 | 11.878 |
| 33 | 31x4; 2x3 | 12.894 | 11.031 | 13.265 | 12.413 |
| 34 | 28x4; 6x3 | 12.894 | 11.085 | 13.265 | 12.413 |
| 35 | 25x4; 10x3 | 12.894 | 11.085 | 12.920 | 12.413 |
| 36 | 22x4; 14x3 | 13.859 | 11.085 | 12.920 | 12.413 |
| 37 | 19x4; 18x3 | 13.859 | 11.608 | 12.920 | 12.317 |
| 38 | 16x4; 22x3 | 13.859 | 11.608 | 12.920 | 12.317 |
| 39 | 13x4; 26x3 | 13.859 | 11.591 | 12.920 | 12.317 |
| 40 | 10x4; 30x3 | 13.842 | 11.591 | 12.920 | 12.317 |
| 41 | 7x4; 34x3 | 13.842 | 11.591 | 12.812 | 12.294 |
| 42 | 4x4; 38x3 | 13.765 | 11.591 | 12.701 | 12.399 |
| 43 | 1x4; 42x3 | 13.765 | 11.591 | 13.041 | 11.401 |
| 44 | 42x3; 2x2 | 15.914 | 11.130 | 12.734 | 12.159 |
| 45 | 40x3; 5x2 | 16.032 | 11.130 | 12.734 | 12.159 |
| 46 | 38x3; 8x2 | 16.391 | 11.130 | 12.734 | 12.159 |
| 47 | 36x3; 11x2 | 16.303 | 11.130 | 12.734 | 12.159 |
| 48 | 34x3; 14x2 | 16.303 | 11.130 | 12.734 | 12.159 |
| 49 | 32x3; 17x2 | 16.288 | 11.130 | 12.734 | 12.159 |
| 50 | 30x3; 20x2 | 16.738 | 11.243 | 12.734 | 12.159 |
| 51 | 28x3; 23x2 | 16.738 | 11.162 | 12.734 | 12.51 |
| 52 | 26x3; 26x2 | 16.738 | 11.162 | 12.734 | 12.518 |
| 53 | 24x3; 29x2 | 16.738 | 11.162 | 12.734 | 12.518 |
| 54 | 22x3; 32x2 | 16.738 | 11.162 | 12.734 | 12.518 |
| 55 | 20x3; 35x2 | 16.738 | 11.162 | 12.734 | 12.518 |
| 56 | 18x3; 38x2 | 16.738 | 11.162 | 12.734 | 12.751 |
| 57 | 16x3; 41x2 | 16.738 | 11.162 | 12.734 | 12.751 |
| 58 | 14x3; 44x2 | 15.690 | 11.162 | 13.478 | 12.751 |
| 59 | 12x3; 47x2 | 15.690 | 11.162 | 13.478 | 12.751 |
| 60 | 10x3; 50x2 | 15.690 | 11.162 | 13.478 | 12.751 |
| 61 | 8x3; 53x2 | 15.690 | 11.162 | 13.478 | 12.751 |
| 62 | 6x3; 56x2 | 15.690 | 11.162 | 13.250 | 12.751 |
| 63 | 4x3; 59x2 | 15.690 | 10.457 | 13.250 | 12.751 |
| 64 | 2x3; 62x2 | 15.690 | 11.861 | 13.250 | 12.751 |
| 65 | 65x2 | 15.690 | 11.861 | 13.467 | 11.000 |

The influence of the group number on the solution value for an input data set DS_E130S274L37A22 is shown in Fig. 4.

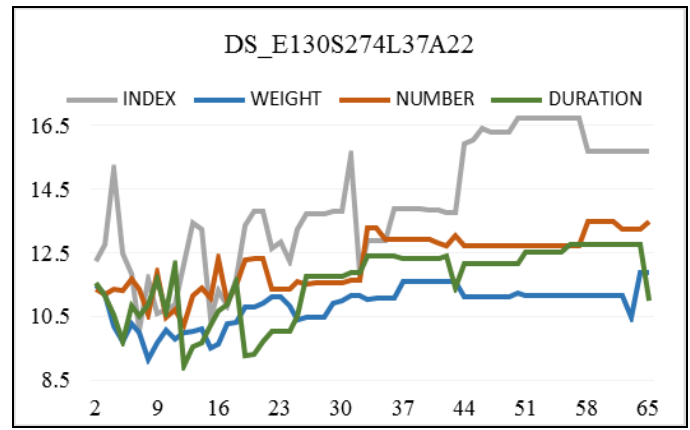

Figure 4.   Influence of the group number (the x axis) on the solution value (the y axis) for DS_E130S274L37A22.

In Fig. V, the best results of the EGB algorithm execution on an input data set DS_E130S274L37A22 (for each sort criteria) are shown.

TABLE V.    THE BEST RESULTS FOR DS_E130S274L37A22

| By | Index | Weight | Number | Duration |
|----|-------|--------|--------|----------|
| Best | m=7: 10.070 | m=8: 9.158 | m=12: 10.239 | m=12: 8.958 |

The influence of the sort criteria on the best solution value for an input data set DS_E130S274L37A22 is shown in Fig. 5.

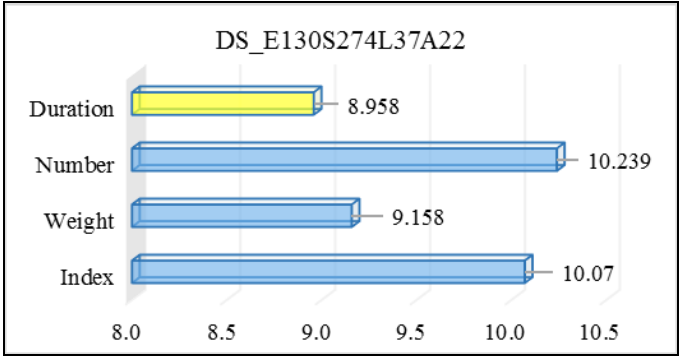

Figure 5.  Influence of the sort criteria (the y axis) on the best solution value (the x axis) for DS_E130S274L37A22.

Fig. IV, V, 4 and 5 show that for the input data set DS_E130S274L37A22 the best found solution is with a value of 8.958. The solution was obtained when the events were sorted by duration and divided into 12 groups (10 groups with 11 events and 2 groups with 10 events). Another good solution (with a value of 9.158) was found when the events were sorted by weight and divided into 8 groups (2 groups with 17 events and 6 groups with 16 events). When the events were sorted by number, the best found solution (with a value of 10.239) is the worst found solution of all other solutions found when sorting the events in the other three criteria.

In Fig. VI, the results of the EGB algorithm execution on input data set DS_E273S549L62A39 are shown.

TABLE VI.    RESULTS FOR DS_E273S549L62A39

| m | Groups | Index | Weight | Number | Duration |
|---|--------|-------|--------|--------|----------|
| 2 | 1x137; 1x136 | 37.582 | 26.480 | 26.072 | 25.406 |
| 3 | 3x91 | 29.974 | 26.323 | 25.494 | 24.452 |
| 4 | 1x69; 3x68 | 34.971 | 24.072 | 23.133 | 23.163 |
| 5 | 3x55; 2x54 | 34.735 | 23.413 | 25.024 | 22.942 |
| 6 | 3x46; 3x45 | 31.980 | **22.068** | 25.073 | 21.861 |
| 7 | 7x39 | 30.382 | 23.387 | 23.183 | 21.745 |
| 8 | 1x35; 7x34 | 28.247 | 23.038 | 25.383 | 21.655 |
| 9 | 3x31; 6x30 | 30.747 | 23.781 | 23.771 | 22.522 |
| 10 | 3x28; 7x27 | 31.623 | 23.125 | 23.608 | 22.341 |
| 11 | 9x25; 2x24 | 27.140 | 22.632 | 24.100 | 23.104 |
| 12 | 9x23; 3x22 | 31.971 | 26.780 | 25.074 | 21.958 |
| 13 | 13x21 | 34.048 | 24.518 | 25.580 | **20.978** |
| 14 | 7x20; 7x19 | 27.902 | 23.419 | 23.987 | 21.707 |
| 15 | 3x19; 12x18 | 28.639 | 23.095 | 25.473 | 21.672 |
| 16 | 1x18; 15x17 | 31.885 | 23.891 | 24.587 | 22.413 |
| 17 | 1x17; 16x16 | 30.846 | 24.603 | **22.948** | 22.800 |
| 18 | 3x16; 15x15 | 30.847 | 24.853 | 23.099 | 22.195 |
| 19 | 7x15; 12x14 | 29.142 | 25.560 | 25.702 | 22.541 |
| 20 | 13x14; 7x13 | **26.562** | 24.867 | 25.303 | 22.536 |
| 21 | 21x13 | 29.473 | 24.310 | 23.600 | 22.953 |
| 22 | 9x13; 13x12 | 31.004 | 23.785 | 24.406 | 22.912 |
| 23 | 20x12; 3x11 | 33.172 | 25.287 | 23.727 | 24.327 |
| 24 | 9x12; 15x11 | 29.104 | 25.891 | 24.332 | 24.325 |
| 25 | 23x11; 2x10 | 32.128 | 25.226 | 25.073 | 23.270 |
| 26 | 13x11; 13x10 | 29.906 | 25.451 | 25.073 | 23.185 |

| 27 | 3x11; 24x10 | 27.928 | 25.506 | 24.942 | 23.032 |
| 28 | 21x10; 7x9 | 30.007 | 23.698 | 24.859 | 23.595 |
| 29 | 12x10; 17x9 | 32.714 | 23.802 | 23.758 | 23.871 |
| 30 | 3x10; 27x9 | 31.514 | 23.410 | 24.629 | 24.150 |
| 31 | 25x9; 6x8 | 35.509 | 24.879 | 25.688 | 23.917 |
| 32 | 17x9; 15x8 | 30.103 | 25.089 | 25.794 | 23.738 |
| 33 | 9x9; 24x8 | 32.117 | 24.983 | 25.867 | 24.795 |
| 34 | 1x9; 33x8 | 32.687 | 25.340 | 25.147 | 24.009 |
| 35 | 28x8; 7x7 | 33.098 | 23.990 | 25.698 | 25.021 |
| 36 | 21x8; 15x7 | 33.433 | 23.906 | 25.933 | 24.727 |
| 37 | 14x8; 23x7 | 36.308 | 24.069 | 26.082 | 25.001 |
| 38 | 7x8; 31x7 | 36.272 | 24.069 | 27.290 | 24.310 |
| 39 | 39x7 | 33.044 | 24.122 | 25.427 | 24.368 |
| 40 | 33x7; 7x6 | 33.046 | 24.120 | 25.404 | 24.368 |
| 41 | 27x7; 14x6 | 33.037 | 23.989 | 25.404 | 24.988 |
| 42 | 21x7; 21x6 | 32.646 | 23.766 | 25.404 | 24.988 |
| 43 | 15x7; 28x6 | 32.748 | 23.669 | 25.287 | 24.988 |
| 44 | 9x7; 35x6 | 33.324 | 23.721 | 25.284 | 25.585 |
| 45 | 3x7; 42x6 | 32.406 | 24.189 | 26.052 | 25.836 |
| 46 ... | 43x6; 2x5 ... | greater than | greater than | greater than | greater than |
| 135 | 3x3; 132x2 | 26.562 | 22.068 | 22.948 | 20.978 |
| 136 | 1x3; 135x2 | 35.686 | 26.363 | 29.085 | 29.026 |

The influence of the group number on the solution value for an input data set DS_E273S549L62A39 is shown in Fig. 6.

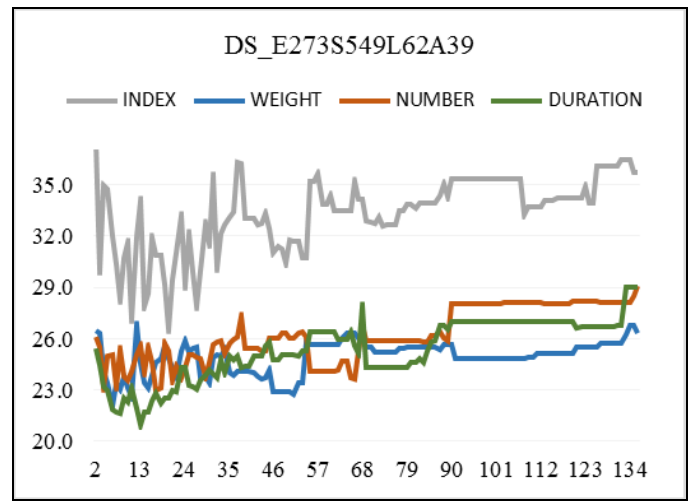

Figure 6.  Influence of the group number (the x axis) on the solution value (the y axis) for DS_E273S549L62A39.

In Fig. VII, the best results of the EGB algorithm execution on an input data set DS_E273S549L62A39 (for each sort criteria) are shown.

TABLE VII.    THE BEST RESULTS FOR DS_E273S549L62A39

| By | Index | Weight | Number | Duration |
|-----|-------|--------|--------|----------|
| Best | m=20: 26.562 | m=6: 22.068 | m=17: 22.948 | m=13: 20.978 |

The influence of the sort criteria on the best solution value for an input data set DS_E273S549L62A39 is shown in Fig. 7.

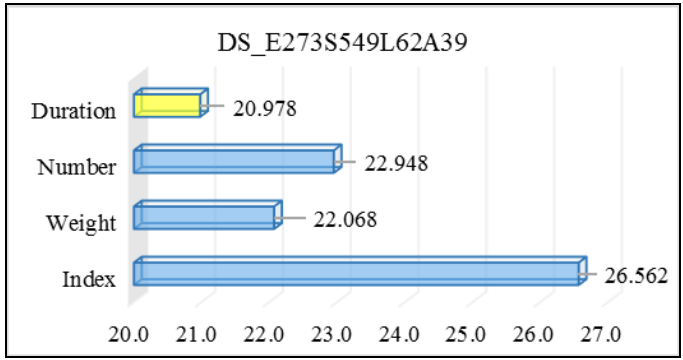

Figure 7.    Influence of the sort criteria (the y axis) on the best solution value (the x axis) for DS_E273S549L62A39.

Fig. VI, VII, 6 and 7 show that for the input data set DS_E273S549L62A39 the best found solution is with a value of 20.978. The solution was obtained when the events were sorted by duration and divided into 13 groups (respectively with 21 events in each). Another good solution (with a value of 22.068) was found when the events were sorted by weight and divided into 6 groups (3 groups with 46 events and 3 groups with 45 events). When the events were sorted by index, the best found solution (with a value of 26.562) is the worst found solution of all other solutions found when sorting the events in the other three criteria.

## IV.    CONCLUSIONS

The best results, the sort criteria and the number of groups after five starts of EGB algorithm (for all input data sets) are shown in Fig. VIII.

TABLE VIII.    THE BEST FIVE RESULTS FOR ALL INPUT DATA SETS

| Input Data Set | Start 1 | Start 2 | Start 3 | Start 4 | Start 5 |
|---|---|---|---|---|---|
| DS_E90S175L29A18 | Weight $m = 3$ 6.530 | Weight $m = 6$ 6.597 | Number $m = 13$ 6.759 | Number $m = 14$ 6.787 | Weight $m = 5$ 6.817 |
| DS_E130S274L37A22 | Duration $m = 12$ 8.958 | Weight $m = 8$ 9.158 | Duration $m = 19$ 9.268 | Duration $m = 20$ 9.307 | Weight $m = 15$ 9.509 |
| DS_E273S549L62A39 | Duration $m = 13$ 20.978 | Duration $m = 8$ 21.655 | Duration $m = 15$ 21.672 | Duration $m = 14$ 21.707 | Duration $m = 7$ 21.745 |

The ratio between the best solutions and the sort criteria (according to number, weight and duration) is shown in Fig. 8.

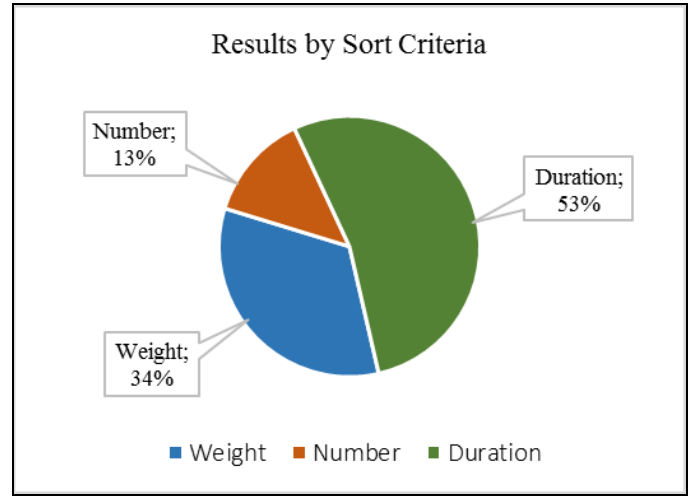

Figure 8.    Ratio between the best solutions and the sort criteria.

Fig. VIII and 8 show that the EGB algorithm found 8 out of 15 best solutions (53%), when the events were sorted by duration. The other 5 solutions (34%) were obtained when the events were sorted by weight. And only 2 solutions (13%) were obtained when the events were sorted by number.

The ranges that contain the groups with the best results for all input data sets are shown in Fig. IX.

TABLE IX.    RANGES OF THE GROUPS WITH THE BEST RESULTS

| Input Data Set | $m$ | Range | Range calculated by m |
|---|---|---|---|
| DS_E90S175L29A18 | 45 | [3, ..., 14] | [$m$ / 15.0, ..., $m$ / 3.2] |
| DS_E130S274L37A22 | 65 | [8, ..., 20] | [$m$ / 8.1, ..., $m$ / 3.25] |
| DS_E273S549L62A39 | 136 | [7, ..., 15] | [$m$ / 19.4, ..., $m$ / 9.1] |

The results obtained show that the range containing all the groups with the best solutions found is [$m$ / 33.3, ..., $m$ / 6.67] (summarized from the results for all input data sets).

After the analysis of the results the following conclusions can be made: 1) the EGB algorithm can be used to solve real UCTP; 2) the number of groups influences on the quality of the solutions found; 3) the number of the tested groups of events can be reduced considering only those that are within the range [$m$ / 33.3, ..., $m$ / 6.67].

The study presented in this paper may be extended in two guidelines: 1) optimization of the EGB algorithm from the point of view of computational complexity and 2) defining more precisely the range of tested groups through conducting additional experiments.

## REFERENCES

[1]    C. Gotlieb, "The construction of class teacher timetables," In C. M. Popplewell, Editor, IFIP Congress 62, pp. 73–77, North-Holland, 1963.

[2]    A. Wren, "Scheduling, timetabling and rostering – a special relationship," The practice and theory of automated timetabling I: Selected papers from 1st International Conference (Patat I), Edinburgh, UK, Lecture notes in computer science 1153, Springer-Verlag, pp. 46-75, 1996.

[3] S. Even, A. Itai, and A. Shamir, "On the complexity of timetable and multi-commodity flow problems," SIAM Journal of computing, vol. 5, no. 4, pp. 691-703, 1976.

[4] D. Werra, "An introduction to timetabling," European journal of operational research, vol. 19, pp. 151–162, 1985.

[5] A. R. Komijan, and M. N. Koupaei, "A mathematical model for university course scheduling: a case study," International Journal of Technical Research and Applications, vol. 19, pp. 20-25, 2015.

[6] R. Chen, and H. Shih, "Solving university course timetabling problems using constriction particle swarm optimization with local search," Algorithms, vol. 6, pp. 227-244, 2013.

[7] M. S. Kohshori, and M. S. Abadeh, "Hybrid genetic algorithms for university course timetabling," International Journal of Computer Science Issues, vol. 9(2), pp. 446-455, 2012.

[8] E. K. Burke, J. Marecek, A. J. Parkes, and H. Rudova, "Decomposition, reformulation, and diving in university course timetabling," Computers & OperationsResearch, vol. 37, pp. 582-597, 2010.

[9] M. Carter, and G. Laporte, "Recent Developments in Practical Course Timetabling," The practice and theory of automated timetabling II: Selected papers from 2nd International conference (PATAT II), Toronto, Canada, Lecture notes in computer science 1408, Springer-Verlag, pp. 3-19, 1998.

[10] E. Burke, K. Jackson, J. Kingston, and R. Weare, "Automated University Timetabling: The State of the Art," The computer journal, 40 (9), pp. 565-571, 1997.

[11] A. Schaerf, "A survey of automated timetabling," Artificial intelligence review, vol. 13 (2), pp. 87-127, 1999.

[12] E. Burke, and S. Petrovic, "Recent research directions in automated timetabling," European journal of operational research, vol. 140 (2), pp. 266-280, 2002.

[13] S. Petrovic, and E. Burke, "University timetabling. handbook of scheduling: algorithms, models and performance analysis," chapter 45, (editor: J. Leung), CRC Press, 2004.

[14] L. Kang, and G. White "A logic approach to the resolution of constraints in timetabling," European journal of operational research, vol. 61, pp. 306-317, 1992.

[15] D. Werra, "Extensions of colouring models for scheduling purposes," European journal of operational research, vol. 92, pp. 474-492, 1996.

[16] D. Costa, "A tabu search for computing an operational timetable", European journal of operational research," vol. 76, pp. 98-110, 1994.

[17] M. Elmohamed, P. Coddington, and G. Fox, "A comparison of annealing techniques for academic course scheduling," The practice and theory of automated timetabling II: Selected papers from 2nd International conference (PATAT II), Toronto, Canada, Lecture notes in computer science 1408, Springer-Verlag, pp. 92-112, 1998.

[18] S. Abdullah, E. Burke, and B. McCollum, "An investigation of a variable neighborhood search approach for course timetabling," The proceedings of the 2nd Multidisciplinary international conference on scheduling: Theory and applications (MISTA 2005), New York, USA, p. 413-427, 2005.

[19] K. Patrick, and Z. Godswill, "Greedy ants colony optimization strategy for solving the curriculum based university course timetabling problem," British Journal of Mathematics & Computer Science, vol. 14(2), pp. 1-10, 2016.

[20] F. Partovi, and B. Arinze, "A knowledge-based approach to the faculty course assignment problem," Socio-economics planning science, vol. 29 (3), pp. 245-256, 1995.

[21] S. E. Soliman, and A. E. Keshk, "Memetic algorithm for solving university course timetabling problem," International Journal of Mechanical Engineering and Information Technology, vol. 3(8), pp. 1476-1486, 2015.

[22] R. Lewis, and B. Paechter, "New Crossover operators for timetabling with evolutionary algorithms," The 5th International conference on recent advances in soft computing (RASC 2004), Nottingham, UK, vol. 5, pp. 189-195, 2004.

[23] V. Kralev, "A model for the university course timetabling problem," International journal "Information technologies & knowledge", vol. 3 (3), pp. 276-289, 2009.

[24] V. Kralev, "A genetic and memetic algorithm for solving the university course timetabling problem," International journal "Information theories & applications", vol. 16 (3), pp. 291-299, 2009.

[25] V. Kralev, and R. Kraleva, "Variable neighborhood search based algorithm for university course timetabling problem," proceedings of the Fifth international scientific conference, FMNS-2013, pp. 202-214, 2013.

[26] V. Kralev, R. Kraleva, and N. Siniagina, "An integrated system for university course timetabling," Proceedings of the third international scientific conference – FMNS2009, vol. 1, pp. 99-105, 2009.

[27] R. P. Badoni, D. K. Gupta, and P. Mishra, "A new hybrid algorithm for university course timetabling problem using events based on groupings of students," Computers & industrial engineering, vol. 78, pp. 12–25, 2014.

[28] E. Falkenauer, "Genetic algorithms for grouping problems," New York: Wiley, 1998.